# Image-Audio Encoding to Improve C**2** Decision-Making in Multi-Domain Environment

Piyush K. Sharma[1] and Adrienne Raglin[2]

Army Research Laboratory[1,2]
Adelphi, MD 20783, USA

**Abstract.** The military is investigating methods to improve communication and agility in its multi-domain operations (MDO). Nascent popularity of Internet of Things (IoT) has gained traction in public and government domains. Its usage in MDO may revolutionize future battlefields and may enable strategic advantage. While this technology offers leverage to military capabilities, it comes with challenges where one is the uncertainty and associated risk. A key question is how can these uncertainties be addressed. Recently published studies proposed information camouflage to transform information from one data domain to another. As this is comparatively a new approach, we investigate challenges of such transformations and how these associated uncertainties can be detected and addressed, specifically unknown-unknowns to improve decision-making.

**Keywords:** IoBT · Multi-Domain Operation · Decision-Making.

## 1 Introduction

### 1.1 Background and Related Work

The modern world is significantly impacted by technology and the dynamics of a globally connected infrastructure. With this new environment comes greater challenges to the process of making decisions in many arenas. Leaders and decision-makers must consider the impact of various factors including those that fall into the category of known and unknown sources of data [9].

While not a new concept the definitions for categorizing known and unknowns has been presented in several papers. When conditions are Known-Knowns: then the condition are where there is knowledge we are aware of and understand, Known-Unknowns: the condition where there is knowledge we are unaware of but do not understand, Unknown-Knowns: the condition where there is knowledge we understand but are unaware of, and Unknown-Unknowns: the condition where there is knowledge we do not understand nor are we aware of [6]. In Figure 1 the discussion of known and unknown partitioning of knowledge is centered around a question. The ones selected in the figure are related to awareness and understanding of risk.

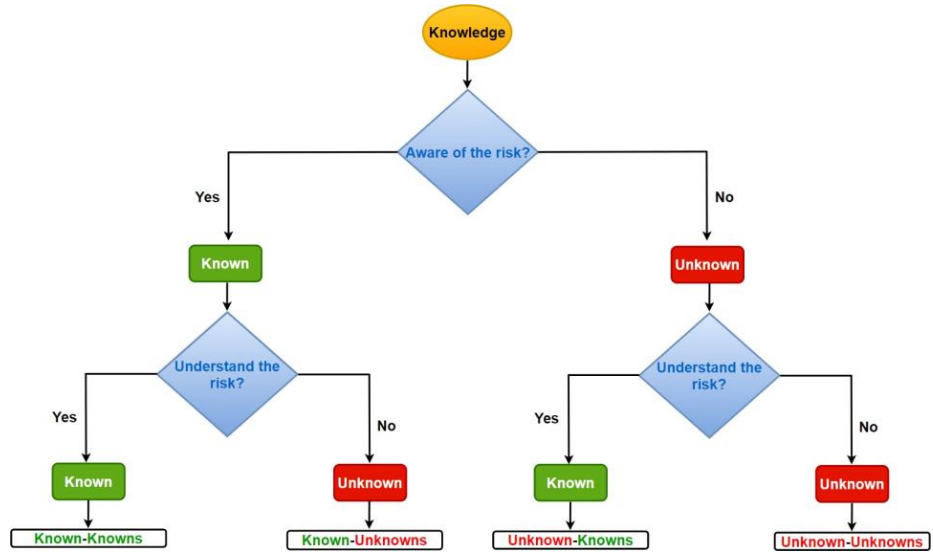

Fig. 1: Flowchart for possible known and unknown pair decision-making.

Out of the four, Known-Knowns is the most obvious one where one can have the complete knowledge of a given problem, whereas Unknown-Unknowns is the completely opposite, and also the most challenging one. Hence, focus should be developing strategies to find likely unknowns so that they convert to knowns pieces of data. However, in many scenarios this may not be trivial, which may require contingency plans and adaptability skills to unforeseen situations.

Known-Unknowns mission plan need to be observed thoroughly. Nevertheless, due to known component, with enough time and resource investment, a plausible can be found. Finally, to deal with Unknown-Knowns [11,22,23], humans are the best known intuitive models with great precognition [5]. Therefore, including suggestions from an individual or group could aid in classifying data that are missed and hence considered unknown by the machine learning model.

We provide a visual representation of aforementioned regions of uncertainty associated with our knowledge about Knowns and Unknowns in Figure 2. In this research we define unknowns as unseen or undetected classes of objects within imagery data that can be uncovered or re-categorized as knowns through the application of the image-audio encoding scheme described in Section 3.1.

### 1.2 Motivation and Challenges

Any decision is heavily influenced by the existence of risk, any process that can help with the identification and understanding of knowns and unknowns is ideal. Moreover, the identification and detection of unknown data could minimize risks. However, it is common to face situations where prior knowledge is not a luxury and only a few data samples are available for analysis. Military decision makers, such as commanders, may have little choice when making crucial decisions, and may ultimately solely depend on their expertise and the influx of new data. They may use their previous experience to analyze incoming information and capture possible unknown data in an effort to minimize the risk. This approach may still not cover all unknowns.

The work in this paper is motivated by the primary challenges in decision-making that we rely solely on the availability of meaningful and sufficient data to support decisions. Also, decision makers must have confidence in the performance and results from techniques used to provide data to support decisions. Hence, we investigate methods to increase the confidence level in decision-making process when the performance of a deep learning model is curbed due to the absence of abundant data samples. We look to how a trained model might be able to detect and identify unknown (undetected) objects with high precision; the model's ability to discriminate whether a new observation belongs to the known or unknown class.

Impetus behind this work arises from a problem within the U.S. Army's IoBT CRA project where devices are, categorized as; red (enemy), gray (neutral), blue (friend) assets. Class attributes and behavior are extremely uncertain, relating to aforementioned known or unknown challenges in Section 1.1, because either data from a friendly source can be compromised or the enemy can potentially be spoofed to resemble friendly data source [1,2,3,4]. Therefore, the classification of these assets with a higher confidence level is a challenging task. An initial step to address this challenge could be taking the data from these devices, for example images, text, or audio and investigating whether unknown pieces of data can be classified as known pieces of data.

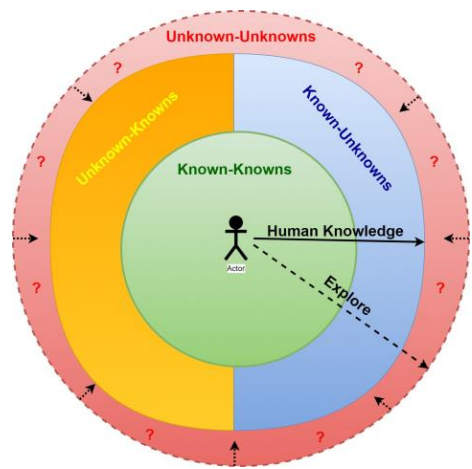

Fig. 2: A visual representation of our proposed approach to illustrate the premises of known and unknown pairs. The frontier of human knowledge become ambiguous and chaotic as we move away from the center outside the green region and step into the other colored regions. The `?' represents region that needs to be explored. Dotted circumference of red region indicates unboundedness of this region due to the lack of any knowledge about this region and its existence. Dotted arrows pointing inwards indicate that the goal should be to converge this red region to any possible yellow, blue or green region. In that order, ideally every encompassing region should be converged to a region it encompasses.

4 P. K. Sharma et al.

### 1.3 Proposed Approach

Our approach includes choosing image data and building a deep learning framework to address the classification challenges. Image classes are specifically chosen to represent landscapes similar to terrains commonly used in military operations.

Thus, our framework consists of two separate components; classification on images obtained from original dataset, and classification on audio signals obtained from images using image-audio encoding scheme (Section 3.1).

Because encoding transforms data from one data domain (image) to other (audio), an information loss is expected. In order to address aforementioned challenges of transformed data samples, we raise the following questions; Can we improve the model's performance when data has been transformed using encoding scheme thus to potentially convert Unknowns to Knowns? How can we compensate for a low model performance so that previously unknown data can be used to improve confidence in decision-making process? What is the trade-off between model's performance and correctly categorizing data to support decision-making?

## 2 Dataset Description

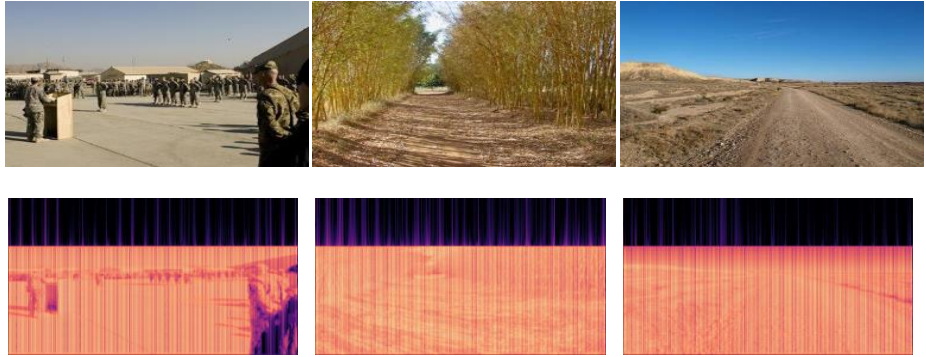

Fig. 3: From left to right. (Above) Sample images from classes: army_base, bamboo_forest, desert_road. Notice similarity in images (specially between bamboo_forest and desert_road ) explaining difficulty in discrimination. (Below) Mel-Spectrograms of corresponding images.

Because our goal is to to investigate how uncovering additional data can support decision making by using encoded data, we evaluate our image-audio encoding approach on a small subset of Places: A 10 million Image Database for Scene Recognition; a large dataset consisting of millions of images from a number of classes [21]. We build a multi-class classification problem for 3 different classes by selecting 2000 training and 100 validation images for each class. Thus, our training and validation datasets consisted of 6000 and 300 images respectively. For camouflage, we transform images to audio signals and collect output in .wav file format and use MFCC to compute audio features. Each audio clip gives 1 MFCC descriptor of 1228 dimensions. Therefore, our train data consisted of a total 6000 instances, each representing a single MFCC descriptor of 1228 dimensions. In order to understand the impact of camouflage; how well the data patterns are preserved, we employ deep leaning approaches on image and audio datasets (before and after encoding) as mentioned in section Section 3.2.

For visual representation of transformed images into audio melodies, we draw their Mel Spectrograms. Whereas a spectrogram plots spectrum of audio frequencies varying with time, Mel Spectrograms plots frequencies on mel scale which perceive evenly spaced to the listeners (Figure 3).

## 3 Experiments

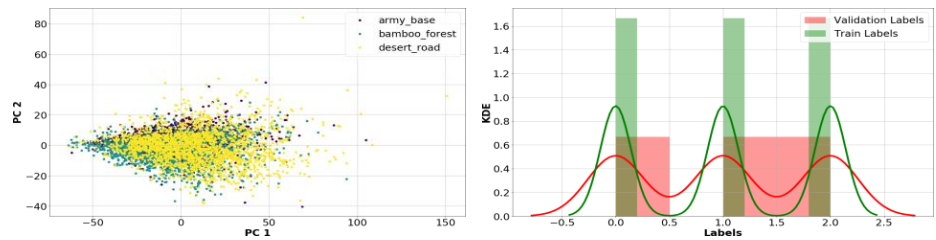

Fig. 4: (Left) We notice challenge in discriminating three classes due to similar variance shown in its PCA scatter plot. (Right) Uniform distribution of class labels shows that ours is a balanced data.

### 3.1 Image-Audio Encoding

The audio signal is a three-dimensional signal representation of time, amplitude and frequency which can be characterized by discriminable features. In determining discriminable features, one analyzes qualitative and quantitative properties of audio. There are many audio feature extraction techniques, such as, zero-crossing rate (ZCR), short-time energy (STE), Linear Prediction Coefficients (LPC), Linear Prediction Cepstrum Coefficients (LPCC), Homomorphic Cepstral Coefficients (HCC), Bark- Frequency Cepstral Coefficients (BFCC), etc. A popular frequency based techniques, Mel-Frequency Cepstral Coefficients (MFCC), has coefficients that collectively make up an MFC (Mel-Frequency Cepstrum) and are derived from a type of cepstral representation of the audio clip. MFC is a representation of the short-term power spectrum of a sound, based on a linear cosine transform of a log power spectrum on a nonlinear mel-scale of frequency [8], [15]. The choice of feature representation impacts classification results. So called good features capture pattern information useful for the classifier to draw distinctions. Beyond feature vector, the distribution that gives rise to audio instantiations also provides useful information for classification.

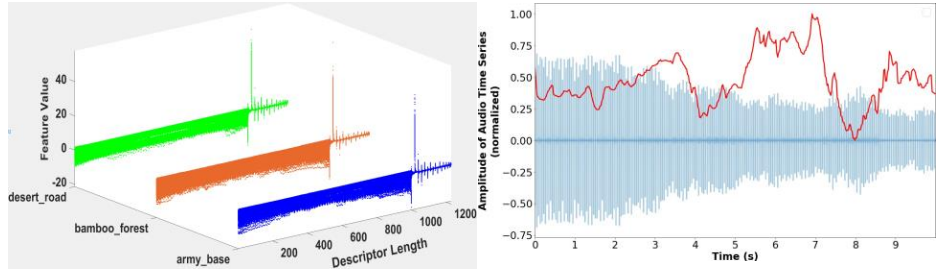

Fig. 5: (Left) 3D visualization of audio data features obtained after image encoding. (Right) Illustration of Spectral Centroid or Center of Mass (in red) computed as the weighted mean of frequencies for an audio signal.

We build a multi-class classification problem with 3 different classes by selecting 2000 training and 100 validation images for each class. Thus, our training and validation datasets consisted of 6000 and 300 images respectively. For camouflage, we transform images to audio signals and collect output in .wav file format using the proposed encoding scheme [18], [19], and use MFCC to compute audio features. Each audio clip gives 1 MFCC descriptor of 1228 dimensions. Therefore, our train data consisted of a total 6000 instances, each representing a single MFCC descriptor of 1228 dimensions (Figure 5). Principal Component Analysis (PCA) is often used to visualize a high dimensional data by projecting it onto a low dimensional subspace [12]. A visualization of audio data with PCA is shown in Figure 4.

### 3.2 Deep Learning

For audio data classification, we build a multi-layered Deep Neural Network (DNN) and tune its hyperparameter searching over a large grid with Batch Size, Epochs, Optimizer, Learn Rate, Momentum, Initial Mode, Activation, Dropout Rate, Weight Constraint, and Neurons.

For image data, we employ transfer learning for which training weights are available after combining the training set of ImageNet 1.2 million data with Places365-Standard 1.8 million data (with at most 5000 images per category) to train VGG16-hybrid1365 model in the Keras framework [13,16]. A summary of the trainable parameters of the model is provided in Table 1. We fine-tune this model by tweaking its parameters for selected dataset and freeze all of the layers except the last 4 convolutional layers which we use for training. Fine-tuning avoids limitations of model by not training from scratch on small data and saving training time (because less parameters will be updated in training).

Table 1: Model summary of trainable parameters

| Layer (type) | Output Shape | Param # |
|---|---|---|
| vgg16-hybrid1365 (Model) | (None, 7, 7, 512) | 14714688 |
| atten_2 (Flatten) | (None, 25088) | 0 |
| dense_3 (Dense) | (None, 1024) | 25691136 |
| dropout_2 (Dropout) | (None, 1024) | 0 |
| dense_4 (Dense) | (None, 3) | 3075 |

Total params: 40, 408, 899
Trainable params: 32, 773, 635
Non-trainable params: 7, 635, 264

### 3.3 Model Performance and Hardware Infrastructure

We evaluate each models' performance by computing its respective validation accuracy and validation loss. A summary of results is provided with tables, confusion matrices, and plots in Section 4. For visualization of the performance measurement, we plot ROC-AUC and Precison-Recall curves. Because we have a multi-class problem, we binarize the output and draw one curve per class label.

For image-audio encoding and audio feature extraction, we used MATLAB. For data preprocessing and deep learning models, we used the latest versions of Python and Keras framework with TensorFlow as a backend [7]. Our computing system consisted of 128 GB RAM for CPU, and NVIDIA Quadro P3200 6144MB - Memory Type: GDDR5 (Samsung). Because deep learning models focus on high accuracy often lacking emphasis on computational complexity, we also provide a run-time comparison.

## 4  Results and Discussion

Here we report empirical validation results for image and audio datasets with aforementioned deep learning approaches. We provide a detailed comparative analysis of models' classification performances before and after camouflage with image-audio encoding scheme. The anti-diagonal of confusion matrices show that in comparison to transfer learning on image data, DNN on audio data had only 3, 4, and 8 misclassifications for labels, army_base, bamboo_forest, and desert_road respectively (Figure 6). This is pretty good when considering the fact that transfer learning had an advantage of using pre-trained weights of a model trained on millions of samples, whereas DNN was trained only on 6000 audio samples.

Run time comparison also suggests that DNN outperforms transfer learning by being approximately 12.5 times faster (Table 2).

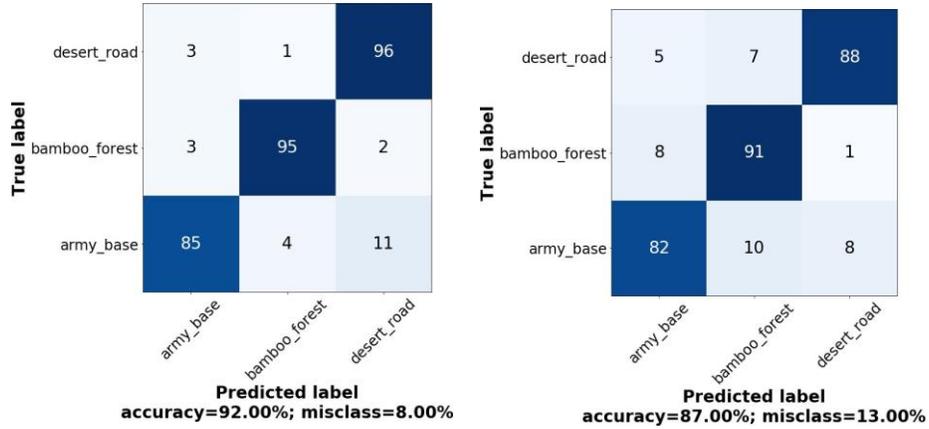

Fig. 6: The anti-diagonal elements of confusion matrices confirm recall values given in Table 3 and Table 4 for (Left) Transfer Learning (Right) DNN.

Table 2: Comparison of models' performance

| Model | Data | Platform | Time taken to run the model (Hours-Minutes-Seconds) |
|---|---|---|---|
| Transfer Learning | Image | GPU | 0 : 14 : 04.589522 |
| DNN | Audio | GPU | 0 : 01 : 17.319555 |

### 4.1   Image Dataset (Before Encoding)

As expected, Transfer Learning performed quite well achieving on average 92% accuracy with 8% loss on validation data (Figure 7). This connotes to our initial assumption that when training data samples are unavailable in abundance, pre-trained models on a big data can leverage model's performance (Section 3.2). Summary of other performance measures is provided in Table 3. The f1-score is the harmonic mean of precision and recall. The support is the number of samples of the true response that lie in that class. As we selected our validation data with 100 instances per class, we see equal number of samples for each class. A quick comparison confirms that these results tally with the confusion matrix and accuracy, ROC-AUC and Precision-Recall curves.

Table 3: Transfer Learning Classification Summary

|              | precision | recall | f1-score | support |
|--------------|-----------|--------|----------|---------|
| army_base    | 0.93      | 0.85   | 0.89     | 100     |
| bamboo_forest| 0.95      | 0.95   | 0.95     | 100     |
| desert_road  | 0.88      | 0.96   | 0.92     | 100     |
| accuracy     |           |        | 0.92     | 300     |
| macro avg    | 0.92      | 0.92   | 0.92     | 300     |
| weighted avg | 0.92      | 0.92   | 0.92     | 300     |

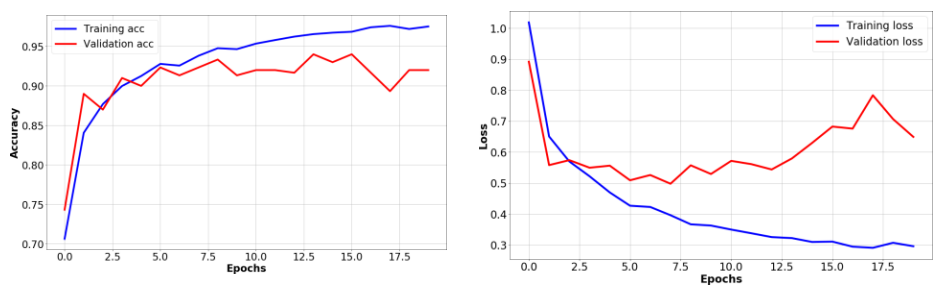

Fig. 7: Transfer Learning on image data. (Left) Training and Validation Accuracy (Right) Training and Validation Loss.

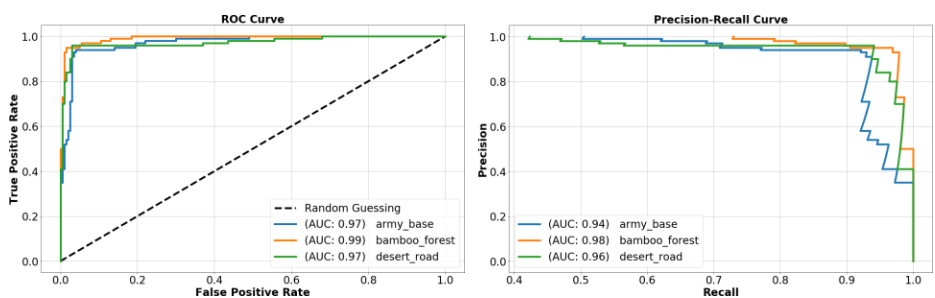

Fig. 8: Transfer Learning on image data. (Left) ROC-AUC (Right) Precision-Recall Curve .

### 4.2 Audio Dataset (After Encoding)

Although, precision achieved on image data (before encoding) is higher for army_base and bamboo_forest than on audio data (after encoding), we notice that it is higher by 3% for desert_road on audio data (Table 3 and Table 4). Our selection of image classes suggests that images with very similar distributions are difficult to discriminate after they are encoded to audio (Figure 3), specifically when there are only a few training samples (Figure 4). However, confidence level in low classification performance can be compensated by focusing on known and unknown realms through direct command involvement in consolidating model's output (accuracy and precision) with decision-maker's intuition (Section 1). Perhaps a reasonable risk can be taken by transmitting a few selected images over the network to command office which can be tallied against audio signal mismatches by DNN. A trade-off between lowering unknown risk and improving confidence in decisions requires training experience and should be at decision-maker's discretion.

Finally, a visual comparison from ROC-AUC and Precision-Recall curves confirms that the deep learning models always results in a greater number of correct decisions (high true positive rate at a very low false positive rate for ROC-AUC, and a high recall without any false positive predictions for Precision-Recall curves) regardless of outcome. AUC shows the probabilities for correctly classifying each class.

Table 4: DNN Classification Summary

|               | precision | recall | f1-score | support |
|---------------|-----------|--------|----------|---------|
| army_base     | 0.86      | 0.82   | 0.84     | 100     |
| bamboo_forest | 0.84      | 0.91   | 0.87     | 100     |
| desert_road   | 0.91      | 0.88   | 0.89     | 100     |
| accuracy      |           |        | 0.87     | 300     |
| macro avg     | 0.87      | 0.87   | 0.87     | 300     |
| weighted avg  | 0.87      | 0.87   | 0.87     | 300     |

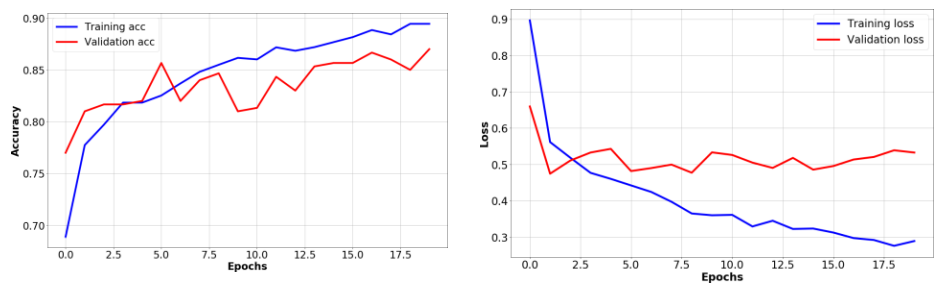

Fig. 9: DNN on audio data. (Left) Training and Validation Accuracy (Right) Training and Validation Loss.

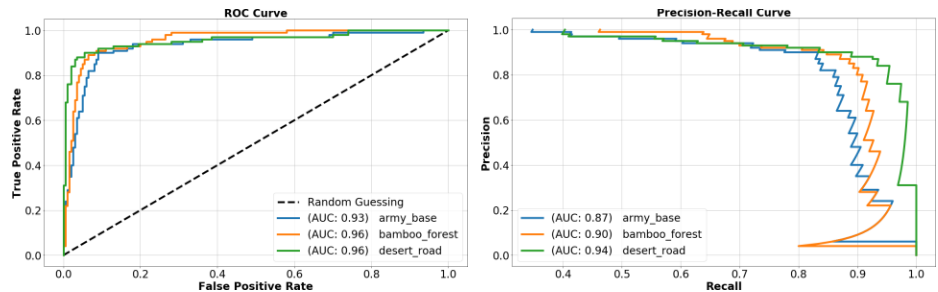

Fig. 10: DNN on audio data. (Left) ROC-AUC (Right) Precision-Recall Curve
.

## 5 Conclusion

In this paper we contributed an approach for utilizing encoded data to recategorize unknown data to known data in order to improve decision-making in Multi-Domain Environments (MDE) where correct detection and classification of data is critical. New methods are needed to help uncover data that can potentially highlight unforeseen risk by reducing uncertainties in categorizing objects more effectively. We explore advancing the techniques for improving accuracy of classifying encoded data and the proposed deep learning models is one approach. This allows the unseen (future) data, which is considered as unknown data, to move to the known data as classification performance improves.

We mentioned a problem within the U.S. Army's IoBT CRA project with three classes, categorized as; red (enemy), gray (neutral), blue (friend) assets. In future work we hope to extend our idea to this problem, for now our analysis involved utilizing a public image dataset with a comparative analysis on datasets before and after encoding.

Our analysis showed that, although, deep learning model's classification performance was better on image data (before encoding) than on audio data (after encoding), its run time complexity was much better on audio data. This perhaps was because transfer learning had an advantage of using pre-trained weights of a model trained on millions of image samples, whereas DNN was trained only on a few audio samples. This connoted to our initial assumption that small data with subtle difference between class distributions can negatively impact model's classification performance.

Although a comparison between confusion matrices showed that DNN on audio data (after encoding) had only a few label mismatched, future work will investigate data regularities and how the data patterns are impacted by image-audio approach.

Moreover, in this work we explored aforementioned categorization of knowns and unknowns with deep learning models, future work will explore other tech-niques which are useful in a similar task with small data, such as, zero-shot learning [14,20], one-shot learning [10], novelty detection [17], etc.